# Yes-Net: An effective Detector Based on Global Information

Liangzhuang Ma   Xin Kan   Qianjiang Xiao   Wenlong Liu   Peiqin Sun
CHENGDU HANDSIGHT INFO-TECH CO
Zone D1004, Tianfu Software Park, Chengdu, Sichuan, China.
ma@handsight.cn

## Abstract

*This paper introduces a new real-time object detection approach named Yes-Net. It realizes the prediction of bounding boxes and class via single neural network like YOLOv2 and SSD, but owns more efficient and outstanding features. It combines local information with global information by adding the RNN architecture as a packed unit in CNN model to form the basic feature extractor. Independent anchor boxes coming from full-dimension k-means is also applied in Yes-Net, it brings better average IOU than grid anchor box. In addition, instead of NMS, Yes-Net uses RNN as a filter to get the final boxes, which is more efficient. For 416 × 416 input, Yes-Net achieves 79.2% mAP on VOC2007 test at 39 FPS on an Nvidia Titan X Pascal.*

## 1. Introduction

As a popular problem in computer vision, object detection is widely discussed and applied in reality. For previous methods that developed based on CNN [1, 2, 3, 4], they got great success, but none of them take global information into account, because of the limited receptive field. It is believed that global information is important to detection task. The prior object detectors, such as YOLOv2 [2], the best intersection-over-union (IOU) between anchor boxes and ground truth boxes is greatly limited by the way the anchors are created. The prior work uses non-maximum suppression (NMS) to predict the output boxes. Obviously, they ignore the relations between adjacent boxes.

We propose a novel base feature extractor named Shun-Net, which has a full image receptive field by using recurrent neural network (RNN). To create better anchor boxes [5], unlike other anchor boxes whose locations are strictly limited in grids, we perform k-means on full data set, and use the final cluster centers as our anchor boxes. Finally, we see the feature of boxes as a sequence and put it into a RNN to select and adjust the outputs. To sum up, we have made it to build a detection network named Yes-net. It is a real-time object detector, and has achieved higher mAP than other current detectors.

## 2. Shun-Net

In Yes-Net, a novel model is introduced as the base feature extractor named Shun-Net. In this model, RNNs and CNN is combined. Similar to Darknet19 [6] that used in YOLO9000 [2] as base architecture, in Shun-Net, general 3×3 filters is adopted and double the number of channels after each pooling step. By adding RNN layers to this model, Shun-Net needs 6.5 billion operations to process a 224 × 224 image and achieves 79.2% top-1 accuracy on ImageNet [9].

### 2.1. Spatial-RNN

In this part, the paper will describe the architecture of RNN layers, specially, on spatial dimension.

Step 1: Treat x-axis as the sequence. Input the feature map row by row. The detail of step1 can be seen in figure1.

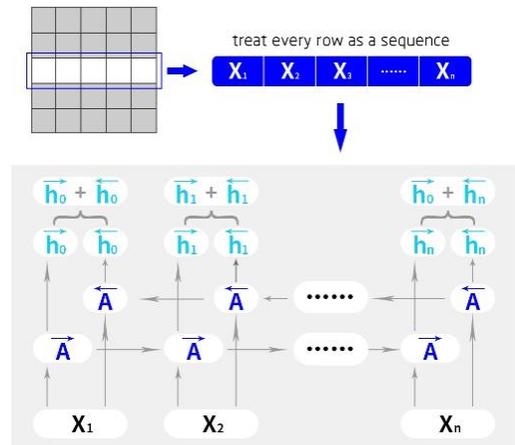

**Figure 1: An explanation of step 1.** (1) Get every row of feature maps from left to right as a sequence and put it to RNN. (2) Get every row of feature maps from right to left as a sequence and put it to RNN. (3) Apply element-wise sum to left-to-right output and right-to-left output, and take the result as the x-axis bi-directional RNN output.

Step 2: Treat y-axis as the sequence, where the input values are the outputs of x-axis Bi-directional RNN. Input the feature map produced by step1 column by column. Detailed steps can be seen in figure 2.

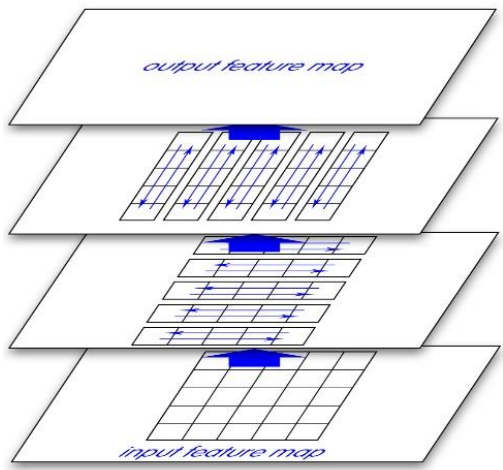

**Figure 2: An explanation of step 2.** (1) Get every column from top to bottom and put it to RNN. (2) Get every column from bottom to top and put it to RNN. (3) Apply element-wise sum to the bi-direction outputs to get the final y-axis output.

### 2.2. Shun-Net with RNN-Packed block

Pack the so-called Spatial-RNN into base units that are RNN Pack-A and RNN Pack-B. Seen in figure 3(a) and figure 3(b), they are similar, and the main difference lies in that one uses concatenation and the other uses element-wise sum. The left branch of the RNN Pack-A is composed by 3 or 5 convolution layers.

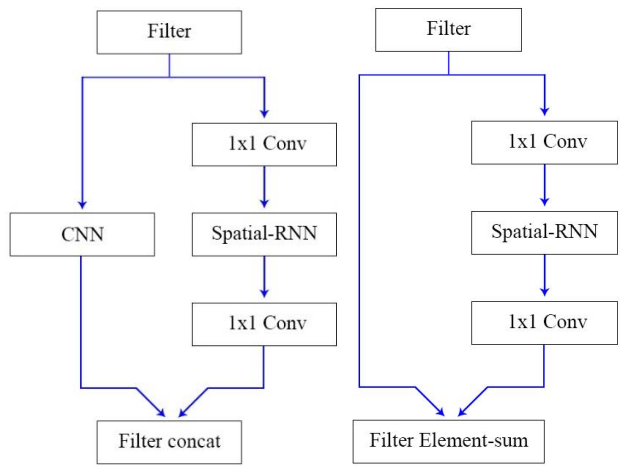

**Figure 3(a): RNNPack-A        Figure 3(b): RNNPack-B**

The model is improved from Darknet19. Respectively insert one RNN Pack-A after the third and the fourth max-pooling layer, and then add an RNN Pack-B and an RNN Pack-A after the last max-pooling layer. The model named Shun-net is described fully in table1.

### 2.3. Training for classification

We train Shun-Net on the standard ImageNet 1000 categories classification dataset [9] for 80 epochs using stochastic gradient descent with momentum (SGDM), with a starting learning rate of 0.05, polynomial rate decay with a power of 4, weight decay of 0.0005 and momentum of 0.9. The batch size is as large as allowed by the GPU memory. We use standard data augmentation tricks including random crops, rotations, scales, noising, sharpen, blur, and hue, saturation, and exposure shifts during training.

| Type | Filters | Size/Stride |
|---|---|---|
| Convolution | 32 | 3 × 3 |
| Maxpool |  | 2 × 2/2 |
| Convolution | 64 | 3 × 3 |
| Maxpool |  | 2 × 2/2 |
| Convolution | 128 | 3 × 3 |
| Convolution | 64 | 1 × 1 |
| Convolution | 128 | 3 × 3 |
| Maxpool |  | 2 × 2/2 |
| RNNPack-A |  |  |
| Convolution | 256 | 1 × 1 |
| Maxpool |  | 2 × 2/2 |
| RNNPack-A |  |  |
| Convolution | 512 | 1 × 1 |
| Maxpool |  | 2 × 2/2 |
| RNNPack-B |  |  |
| RNNPack-A |  |  |
| Convolution | 1024 | 1 × 1 |
| Convolution | 1001 | 1 × 1 |
| Avgpool |  | Global |
| Softmax |  |  |

**Table 1: Shun-Net**

## 3. Anchor box

The concept of anchor box was first presented in Faster R-CNN [5]. Faster R-CNN selects 9 anchor boxes by hand from every cell in the last feature map. In YOLOv2, Joseph Redmon et al. adopt a better method to select anchor boxes. They run k-means on training set to get m cluster centroids to represent the best width and height of anchor boxes in each cell. Compared with YOLOv2, Yes-Net achieves significant improvement to get better anchor boxes.

## 3.1. Problems analysis

In YOLOv2, the IOUs between the anchor boxes of adjacent cells are too high; it means that the information in these boxes is almost the same. Similar information indicates poorer performance of the whole network.

## 3.2. K-means

Instead of selecting 5 anchor boxes in each cell of the last feature map, Yes-Net uses k-means to cluster N anchor boxes based on the whole training set. By this approach, fixed number of anchor boxes which have fix shape and fix center in each grid cell is avoided. In order to include more information in every anchor boxes, a threshold in k-means is implemented. Once IOU between any two cluster centroids is higher than the threshold, it will be merged to create a new one. The distance metric used in k-means is defined as follows:

$$d(box, centroid) = 1 - IOU(box, centroid) \quad (1)$$

## 3.3. Network architecture

Use N anchor boxes selected by k-means may cause some changes in network architecture compared to YOLOv2. Based on the last feature map in YOLOv2, Yes-Net adds two reshaped operations, one is full of connecting layer and another one is made of convolution layer whose filter is $1 \times 1$. After the last convolution layer, N output is got. The details of our network is shown in figure4.

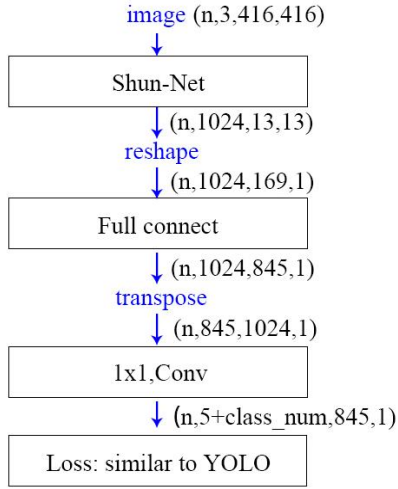

**Figure 4: The Network architecture.** Our network takes n input images, each image is $416 \times 416$, and has three channels. The images will go through a feature extractor which is called Shun-Net. The output of feature extractor will be reshaped, from (n, 1024, 13, 13) to (n, 1024, 169, 1). And the output as an input of a full connect layer will get (n, 1024, 845, 1). Similarly, (n, 1024, 845, 1) will be transposed to (n, 845, 1024, 1). Then, use a convolution layer with $1 \times 1$ filter to get outputs.

## 3.4. Loss Function

The loss function of Shun-Net has only a little difference from YOLOv2 in form, but there is a significant variance in physical meaning:

$$\lambda_{coord} \sum_{i=0}^{N} \mathbb{1}_i^{obj}[(x_i - \hat{x}_i)^2 + (y_i - \hat{y}_i)^2]$$
$$+\lambda_{coord} \sum_{i=0}^{N} \mathbb{1}_i^{obj}\left[(\sqrt{w_i} - \sqrt{\hat{w}_i}) + (\sqrt{h_i} - \sqrt{\hat{h}_i})\right]$$
$$+\sum_{i=0}^{N} \mathbb{1}_i^{obj}(C_i - \hat{C}_i)^2$$
$$+\lambda_{noobj} \sum_{i=0}^{N} \mathbb{1}_i^{noobj}(C_i - \hat{C}_i)^2$$
$$+\sum_{i=0}^{N} \mathbb{1}_i^{obj}(p_i(c) - \hat{p}_i(c))^2 \quad (2)$$

Where N is the number of anchor boxes, $\mathbb{1}_i^{obj}$ denotes that the $i$th anchor box is "responsible" for that prediction. Compared to YOLO and YOLOv2, it's more flexible. Because the loss function does not require that the anchor box must be in fixed cell, we can choose the anchor box from all anchor boxes in a full image with the highest current IOU with the ground truth.

## 4. Selecting the output boxes

NMS is a general algorithm in object detection used for choosing the final object box. But there are two weaknesses in the algorithm, which lower many detectors' performance. In this paper, we propose a novel method that use a RNN instead of NMS to select the output boxes, which finally improved the generalization ability of our detector. We argue that this method also can be adopted by other detectors. The new network structure is shown in figure5.

### 4.1. Limits of NMS

NMS is often seen in many detectors, like R-CNN [1], YOLO [6] and so on. But, NMS is not a perfect approach to handle output boxes selection. There are two shortcomings in NMS. Firstly, if using NMS in detector design, a threshold must be artificially determined to remove similar boxes. It is quite difficult to select all output boxes rightly from various objects in a fixed threshold. And if the threshold is not rightly chosen, the precision of detector will be decreased. Secondly, when a detector uses NMS, it assumes that these output boxes are independent. While in reality, no matter the boxes from the same class or not, there has great possibility that they share some logical relationship.

## 4.2. Why introduce RNN

In order to make our detector own an ability to take boxes that are close into account when selecting the final output box of an object, the Yes-Net introduces RNN. Using a RNN to replace NMS in detector design, you can overcome the problems of NMS.

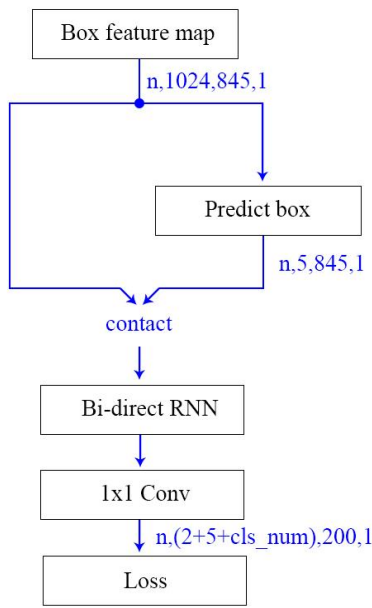

**Figure 5: The Network architecture.** By using feature extractor, feature map is got. Use feature map as the input of predict box layer to get predict boxes. Then concatenate on the predict box and the feature vector for each anchor boxes. These new vectors are sorted according to the value of confidence of each box. Next, select the top-N feature vectors to form a sequence as the input of bi-direct RNN.

## 4.3. Structure a RNN sequence

Concatenate the feature vectors that extracted by shun-net and predict boxes which are the output of the net that are described in section 3.3 to get a new feature vector. Every predict boxes include 5 values. They are x (center of x), y (center of y), w (box width), h (box height) and c (confidence). Then, sort all feature vectors according to the confidence value of each box. And select the top-N feature vectors to form a sequence. Finally, organize the sequence into a shape like Gaussian shape according to the level of confidence of each box. Because higher the value, the more possibility that the confidence box is the answer, the answer box is in the middle of the sequence and it consider other boxes to get a better result. The details of how to organize the sequence is shown in figure6.

## 4.4. Structure a better input of RNN

Based on the feature vector that described in section 4.3, a better input of RNN is structured by adding x2, y2, xy, w2, h2, wh. The x, y, w, h is from the output of the net that described in section 4.3. Because in our detector, the RNN which followed the base feature extractor is shallow. So we artificially expanded the RNN function set by adding x2, y2, xy, w2, h2, wh in input. These new information can effectively help RNN to learn some useful knowledge about the relation between boxes.

## 4.5. Loss Function

During training we optimize the following, multi-part loss function:

$$\frac{1}{K}\sum_{i=0}^{N}\mathbb{1}_i^{obj}[\,(x_i-\hat{x}_i)^2+(y_i-\hat{y}_i)^2$$
$$+\left(\sqrt{w_i}-\sqrt{\hat{w}_i}\right)^2+\left(\sqrt{h_i}-\sqrt{\hat{h}_i}\right)^2+(C_i-1)^2]$$
$$+\sum_{i=0}^{N}\frac{\mathbb{1}_i^{noobj}(C_i)^2}{N-K}-\sum_{i=0}^{N}\mathbb{1}_i^{obj}\frac{\log(S_1)}{K}$$
$$-\sum_{i=0}^{N}\mathbb{1}_i^{obj}\frac{\log(S_0)}{N-K}-\sum_{i=0}^{N}\mathbb{1}_i^{obj}\log(P_i(c)) \qquad (3)$$

In an image, where K denotes there are K objects. N denotes the numbers of predict results for an image. Every result has 3 groups of dimensions which have sizes of 2, 5 and class-number. The first part has 2 dimensions which indicate that there are two likelihoods. One is the possibility that this result is an output, the other one is the possibility that this result is not an output. 5 represents a box includes x, y, w, h, confidence.

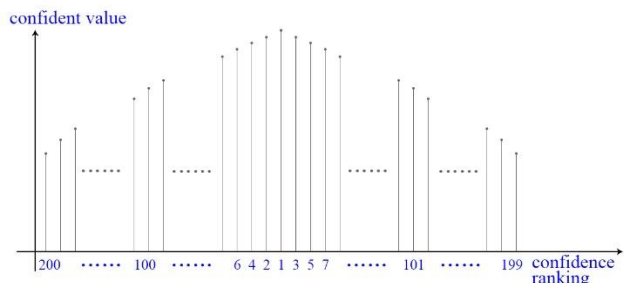

**Figure 6: A RNN sequence.** Select top-N, for example N=200 to form a sequence. In the middle of RNN sequence, the box has maximum confidence value. On both ends of RNN sequence, these boxes have minimized confident value.

## 5. Yes-Net

This chapter will first describe the architecture of Yes-Net which combines all net and can be explained in chapter2,

| Detection Frameworks | Train | mAP | FPS |
|---|---|---|---|
| Fast R-CNN | 2007+2012 | 70.0 | 0.5 |
| Faster R-CNN VGG-16 | 2007+2012 | 73.2 | 7 |
| Faster R-CNN ResNet | 2007+2012 | 76.4 | 5 |
| YOLO | 2007+2012 | 63.4 | 45 |
| SSD300 | 2007+2012 | 74.3 | 46 |
| SSD500 | 2007+2012 | 76.8 | 19 |
| YOLOv2 416 | 2007+2012 | 76.8 | 67 |
| Yes-net 416 | 2007+2012 | 79.2 | 39 |

**Table 2: Results on PASCAL VOC 2012**

chapter3, and chapter4. The training method of Yes-Net will be demonstrated in details. In the end, the experiments will be tested on VOC 2012[10] of Yes-Net.

### 5.1. Architecture

Shun-Net is modified for detection by removing the last convolution layer, avgpool layer, softmax layer and adding some other layers, and two loss layer is also added to the end, seen figure 7. The architecture of Yes-net is designed as two detectors and they both have feedback. The right detector is the same as YOLOv2 aimed at assisted training. And the left detector is designed for getting the final predicted box, there is no need to use non-maximum suppression.

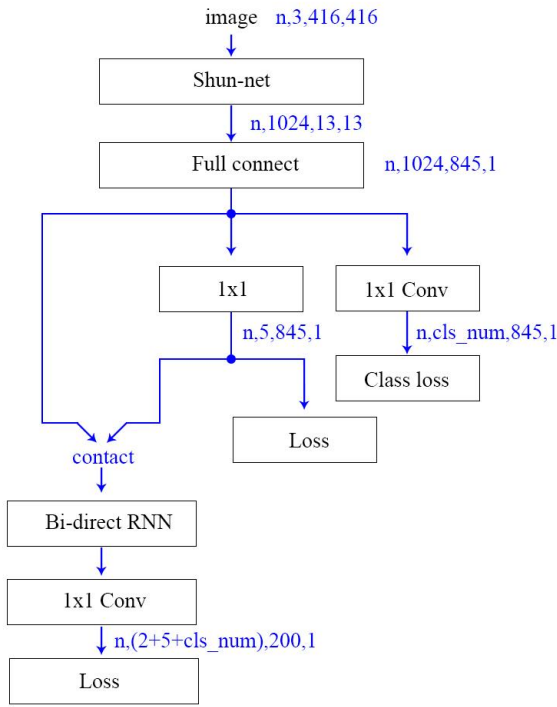

**Figure 7: The Architecture of Yes-Net.**

### 5.2. Training

The overall training can be split into two stages. In the first stage, train the network before YOLO loss layer. The loss function consists of box loss part and class loss part. In this loss function, different weights are given respectively. In the second stage, train Yes-Net according the loss function which be defined in (3), named Yes_loss. There is a trick should be noticing, the feedback of Yes_loss propagate back to fully connect layer directly, there is no need to pass the 1×1 convolution layer.

### 5.3. Experiment Results

Test Yes-net for detection on PASCAL VOC 2012 data set. Table3 shows the comparative performance of Yes-net versus other state-of-the-art detection systems. Yes-net achieves 79.2% mAP and running fast enough for than competing methods.

## 6. Conclusion

This paper introduces Yes-net, an innovative and more comprehensive model. The model is simply constructed and combining CNN with RNN. The model can get the final predict boxes directly and effectively. Unlike other object detection model, Yes-net doesn't need to used non-maximum suppression that can influence the final predict result seriously. Yes-Net adopts RNN instead of NMS to predict the final box better.

Another innovation of Yes-Net is that the model takes global information into account, the box-RNN is designed for extracting the global information of the whole picture. And what's more, Box-RNN used on spatial dimension can get the connections between objects in one picture. For example, to predict a pair of glasses on someone's face, the probability by the model will raise from original 0.5 to 0.9 for it takes connections between objects into account. It is just like the way of human thinking.

# References

[1] Girshick R, Donahue J, Darrell T, et al. Rich Feature Hierarchies for Accurate Object Detection and Semantic Segmentation[C]. Computer Vision and Pattern Recognition. IEEE, 2014:580-587.

[2] Redmon J, Farhadi A. YOLO9000: Better, Faster, Stronger. arXiv preprint arxiv:1612.08242v1, 2016.9

[3] Liu W, Anguelov D, Erhan D, et al. SSD: Single Shot MultiBox Detector. arXiv preprint arxiv:1512.02325v5 2016.9

[4] Dai J, Li Y, He K, et al. R-FCN: Object Detection via Region-based Fully Convolutional Networks. arXiv preprint arxiv:1605.06409v2, 2016.6

[5] Ren S, He K, Girshick R, et al. Faster R-CNN: Towards Real-Time Object Detection with Region Proposal Networks. [C]. International Conference on Neural Information Processing Systems. MIT Press, 2015:91-99.

[6] Redmon J, Divvala S, Girshick R, et al. You Only Look Once: Unified, Real-Time Object Detection[C]. Computer Vision and Pattern Recognition. IEEE, 2016:779-788.

[7] Cho K, Merrienboer B V, Gulcehre C, et al. Learning Phrase Representations using RNN Encoder-Decoder for Statistical Machine Translation[J]. Computer Science, 2014.

[8] Hochreiter S, Schmidhuber J. Long short-term memory. [J]. Neural Computation, 1997, 9(8):1735.

[9] Russakovsky O, Deng J, Su H, et al. ImageNet Large Scale Visual Recognition Challenge. [J]. International Journal of Computer Vision, 2015, 115(3):211-252.

[10] The PASCAL Visual Object Classes Challenge 2012 (VOC2012) Results.